\documentclass{bmvc2k}

\title{BiSeg: Simultaneous Instance Segmentation and Semantic Segmentation with Fully Convolutional Networks}

\addauthor{Viet-Quoc Pham}{quocviet.pham@toshiba.co.jp}{1}
\addauthor{Satoshi Ito}{satoshi13.ito@toshiba.co.jp}{1}
\addauthor{Tatsuo Kozakaya}{tatsuo.kozakaya@toshiba.co.jp}{1}

\addinstitution{
 Corporate Research and Development Center, Toshiba Corporation\\
 1, Komukai-Toshiba-cho, Saiwai-ku, Kawasaki, 212-8582, Japan
}

\runninghead{Pham, Ito, Kozakaya}{BiSeg}

\def\eg{\emph{e.g}\bmvaOneDot}

\def\etal{\emph{et al}\bmvaOneDot}

\def\figref#1{Fig.~\ref{fig:#1}}
\def\tabref#1{Tab.~\ref{tab:#1}}
\def\eqnref#1{(\ref{eq:#1})}

\begin{document}

\maketitle

\begin{abstract}
We present a simple and effective framework for simultaneous semantic segmentation and instance segmentation with Fully Convolutional Networks (FCNs). The method, called BiSeg, predicts instance segmentation as a posterior in Bayesian inference, where semantic segmentation is used as a prior. We extend the idea of position-sensitive score maps used in recent methods to a fusion of multiple score maps at different scales and partition modes, and adopt it as a robust likelihood for instance segmentation inference. As both Bayesian inference and map fusion are performed per pixel, BiSeg is a fully convolutional end-to-end solution that inherits all the advantages of FCNs. We demonstrate state-of-the-art instance segmentation accuracy on PASCAL VOC.
\end{abstract}

\section{Introduction}
\label{sec:intro}
\begin{figure*}
 \centering
 \includegraphics[width=\textwidth]{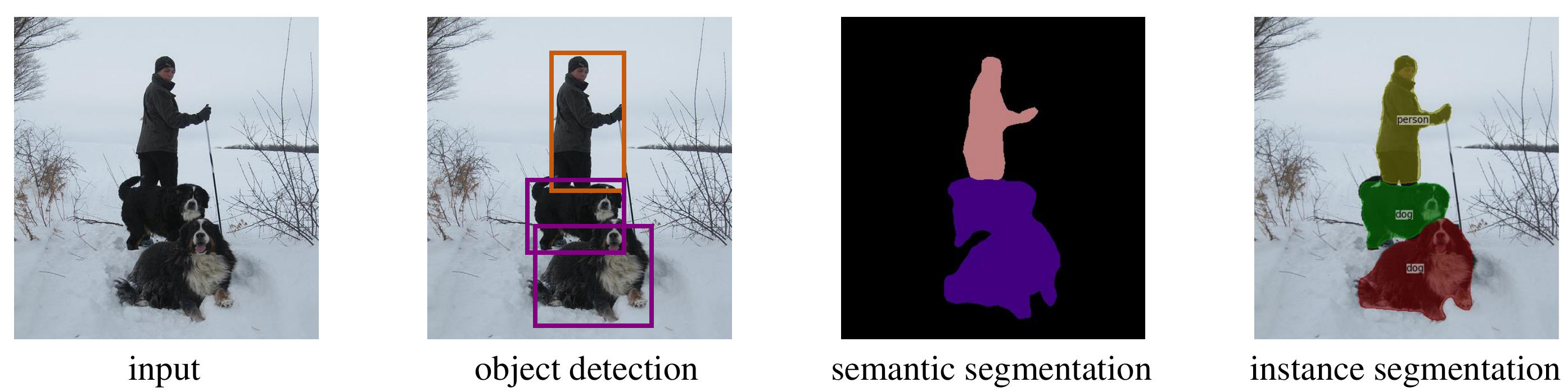}
 \caption{Three important tasks in image recognition: object detection, semantic segmentation and instance segmentation. Instance segmentation requires both high object detection accuracy and precise segmentation.} 
 \label{fig:inst-seg}
\end{figure*}

Object detection and semantic segmentation are problems that have been intensively studied in the field of computer vision, and the accuracy of the proposed solutions has improved rapidly owing to the progress of deep learning \cite{He2016, Chen2016a, Dai2016c}. In semantic segmentation, the task is to classify each pixel into a fixed set of categories without distinguishing among object instances. While on the other hand, object detection detects object instances at the bounding-box level. In this paper, we address the problem of instance segmentation, which aims to localize objects at the pixel level. Instance segmentation is challenging because it requires both high object detection accuracy and precise segmentation (see \figref{inst-seg}).

Early methods \cite{Girshick2014, Hariharan2014, Hariharan2015, Dai2015} address instance segmentation using convolutional neural networks (CNNs), but require mask proposals from external modules \cite{Uijlings2013, Arbelaez2014}. Different from these methods, DeepMask \cite{Pinheiro2015}, SharpMask \cite{Pinheiro2016} and Instance FCN \cite{Dai2016b} learn to propose segment candidates. 

DeepMask \cite{Pinheiro2015} is trained with two objectives: predicting a class-agnostic segmentation mask from a given image patch, and predicting how likely the patch is to contain an object. At test time, DeepMask is applied densely to an image and generates a set of object masks, and corresponding objectness scores. SharpMask \cite{Pinheiro2016} improves DeepMask by generating higher quality masks using an additional top-down refinement step. Its idea is to first generate a coarse mask in a feedforward pass, then refine this mask in a top-down pass using features at successively lower layers.

Instance FCN is a fully convolutional approach for instance mask proposal generation \cite{Dai2016b}. It extends the translation invariant score maps in conventional FCNs to position-sensitive score maps, which are somewhat translation-variant. On top of these instance-sensitive score maps, an assembling module is able to output instance candidate at each position.The concept of position-sensitive score maps will be discussed in details in Section \ref{sec:fcis}.

The common problem of the above three methods \cite{Pinheiro2015, Pinheiro2016, Dai2016b} is that they are only used for mask proposal generation, which is blind to semantic categories and requires a downstream network for category classification. In their implementation, the proposed segment candidates are classified by Fast R-CNN \cite{Girshick2015} in a different process. As these methods adopt a time-consuming image pyramid scanning to find instances at different scales, they are slow at inference time and less accurate.

In a different approach, Dai \etal \cite{Dai2016a} propose an end-to-end solution to address instance segmentation. Their multi-task network cascade (MNC) model consists of three stages: proposing box-level instances, regressing mask-level instances, and categorizing each instance. In the first stage, object instances are proposed in the form of bounding boxes, which are class-agnostic, and are predicted with an objectness score. The second stage takes the shared convolutional features and stage-1 boxes as input, and outputs a pixel-level segmentation mask for each box proposal. The third stage takes the predicted boxes and masks from the previous stages as input, and outputs category scores for each instance.

Recently, Li \etal \cite{Li2017} propose a fully convolutional end-to-end solution for instance segmentation, named FCIS, by extending the idea of position-sensitive score maps in \cite{Dai2016b}. The details will be stated in Section \ref{sec:fcis}. The advantage of this method is that the underlying convolutional representation and the score maps are fully shared for the mask prediction and classification sub-tasks, via a joint formulation with no extra parameters. Furthermore, it operates on box proposals instead of sliding windows, enjoying the recent advances in object detection \cite{Ren2015}. FCIS achieved state-of-the-art performance in both PASCAL VOC \cite{Everingham2010} and MS COCO \cite{Lin2014}. It is also $6 \times$ faster than MNC \cite{Dai2016a}.

In this paper, we present a simple and effective framework for simultaneous semantic segmentation and instance segmentation with Fully Convolutional Networks (FCNs). The method, called BiSeg, predicts instance segmentation as a posterior in Bayesian inference, where semantic segmentation is used as a prior. We extend the idea of position-sensitive score maps used in \cite{Dai2016b, Li2017} to a fusion of multiple score maps at different scales and partition modes, and adopt it as a robust likelihood for instance segmentation inference.

There are advantages to combining semantic segmentation into the instance segmentation framework. First, semantic segmentation can be used to enhance instance segmentation owing to the strong correlation between the two tasks. In related work, Shrivastava and Gupta \cite{Shrivastava2016b} add segmentation to Faster R-CNN \cite{Ren2015} as a complementary task and use it to provide top-down information to guide region proposal generation and object detection. The intuition is that semantic segmentation captures contextual relationships between objects, and will essentially guild the region proposal module to focus attention in the right areas and learn detectors from them. The second advantage is that semantic segmentation can deal with stuff. Different from objects, stuff \cite{Mottaghi2014} (\eg, sky, grass, water) is usually treated as the context in the image. Stuff mostly exhibits as colors or textures and has less well-defined shapes. It is thus inappropriate to use a single rectangular box or a single segment to represent stuff. Combination of object detection from instance segmentation and stuff detection from semantic segmentation is necessary for robust image recognition systems.

Finally, we verify the performance of our method on the PASCAL VOC dataset. BiSeg is a fully convolutional end-to-end solution that inherits all the advantages of FCNs. We achieve the best results, 67.3\% and 54.4\% which are nearly 2\% higher than those of FCIS \cite{Li2017}.

\section{Preliminaries: FCIS}
\label{sec:fcis}
FCIS \cite{Li2017} is the first fully convolutional end-to-end solution for instance segmentation. It performs instance mask prediction and classification jointly by extending the ideas concerning position-sensitive score maps.

\begin{figure*}
 \centering
 \includegraphics[width=\textwidth]{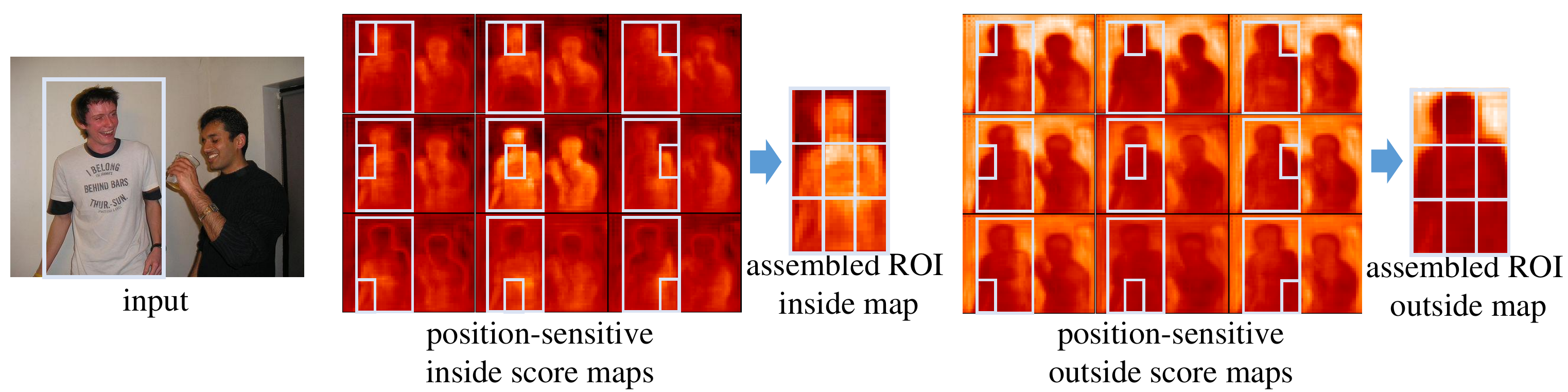}
 \caption{Inside and outside position-sensitive score maps with $3 \times 3$ partitions. The inside/outside likelihood map for each ROI is produced by assembling its $k \times k$ partitions from the corresponding inside/outside score maps.} 
 \label{fig:ps-scoremap}
\end{figure*}

{\noindent \bf Inside/Outside Position-Sensitive Score Maps:} In FCN \cite{Long2015}, the method predicts $C + 1$ score maps representing the likelihood of each category at each pixel. Here, $C$ is the number of object categories, 1 is for background. Instance FCN \cite{Dai2016b} uses $k^2$ position-sensitive score maps to encode the position information with respect to a relative spatial position from $k \times k$ partitions (\eg, the top left part of an object). Based on the above ideas, FCIS \cite{Li2017} proposes a larger set of $2k^2 \times (C+1)$ position-sensitive score maps to encode more detailed semantic information at each pixel. For each object category, each pixel has two scores in each partition, representing the likelihoods of being inside (or outside) the object boundary at a relative position (from $k \times k$ partitions). The score maps are illustrated in \figref{ps-scoremap}.

{\noindent \bf Joint Mask Prediction and Classification:} For each region-of-interest (ROI), the method produces an inside/outside likelihood map by assembling its $k \times k$ partitions from the corresponding inside/outside score maps. On top of the two likelihood maps, the method applies a softmax operation to produce the foreground probability, and a max operation to produce the per-pixel likelihood of the object category, which is then used to infer the classification score by average pooling.

All the per-ROI components are parameter free. The score maps are computed by FCN, and does not involve any feature warping, resizing or fc layers. 

\section{Proposed Approach}
\label{sec:proposed-approach}
\begin{figure*}
 \centering
 \includegraphics[width=\textwidth]{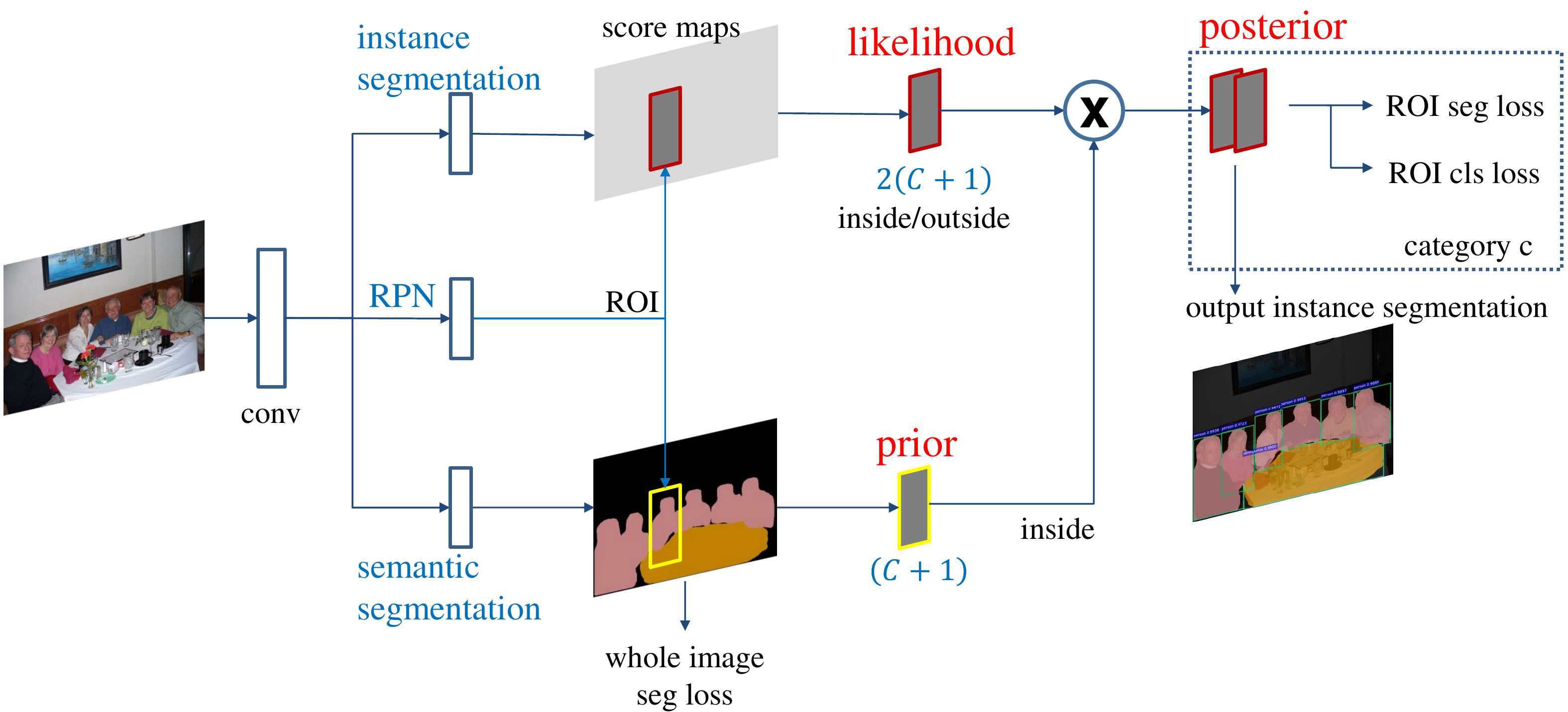}
 \caption{Network architecture of BiSeg. It takes an image of arbitrary size as input and outputs both semantic segmentation and instance segmentation results. RPN, semantic segmentation and instance segmentation sub-networks share the convolutional feature maps. The learnable weight layers are fully convolutional and computed on the whole image. For simplification, we do not show the bounding box regression sub-network here.} 
 \label{fig:biseg-network}
\end{figure*}

In our BiSeg model, the network inputs an image of arbitrary size and outputs both semantic segmentation and instance segmentation results. The network has several sub-networks that share convolutional features: region proposal, bounding box regression, semantic segmentation and instance segmentation. The region proposal sub-network generates ROIs, following the work of Region Proposal Networks (RPN) \cite{Ren2015}. The bounding box regression sub-network refines the initial input ROIs to produce more accurate detection results, as in \cite{Girshick2015}. The semantic segmentation sub-network infers the semantic segmentation probabilities. The instance segmentation sub-network estimates the likelihoods of instance segmentation. Finally, the semantic segmentation and instance segmentation sub-networks are combined to infer the instance segmentation probabilities. Our model is illustrated in \figref{biseg-network}.

{\noindent \bf Network Architecture:} We use the ResNet model \cite{He2016} as the convolutional backbone architecture for feature extraction. We add a $1 \times 1$ convolutional layer on top of the feature map to reduce the dimension from 2048 to 1024. 

Following FCIS \cite{Li2017}, we apply the ``hole algorithm'' \cite{Chen2015} to reduce the feature stride of conv5 layers from 32 to 16. The RPN sub-network, which is fully convolutional, is added on top of the conv4 layers, as in \cite{Dai2016a, Li2017}.

The semantic segmentation sub-network in added on top of the conv5 feature maps. Our implementation is based on the FCN architecture in \cite{Long2015}. This sub-network produces $C+1$ score maps $S$ indicating the segmentation probabilities of $C+1$ categories. To verify the effectiveness of Bayesian inference in our model, we try to keep the model of semantic segmentation as simple as possible. In this paper, we neither consider Conditional Random Field (CRF) \cite{Chen2015, Zheng2015, Schwing2015, Lin2016}, higher order potential \cite{Arnab2016b}, nor domain transform \cite{Chen2016b}. The effect of higher semantic segmentation quality on the instance segmentation performance is an interesting topic that we intend to address in future work.

{\noindent \bf Fusion of Position-Sensitive Score Maps:} In the instance segmentation sub-network, we aim to infer instance segmentation likelihoods $L$ for each ROI. We extend the idea of inside/outside position-sensitive score maps in FCIS to a fusion of multiple score maps at different scales and partition modes. Although it was shown in \cite{Dai2016b} that $7 \times 7$ partitions gave the best overall accuracy, we found that other partition modes worked better in some cases. To reduce the dependence on the partition parameter, we make a fusion of multiple score maps with different partition parameters. Furthermore, to enlarge the variation of score maps, we also generate score maps from different scales, and employ the skip architecture in \cite{Long2015} to perform the map fusion.

\begin{figure*}
 \centering
 \includegraphics[width=\textwidth]{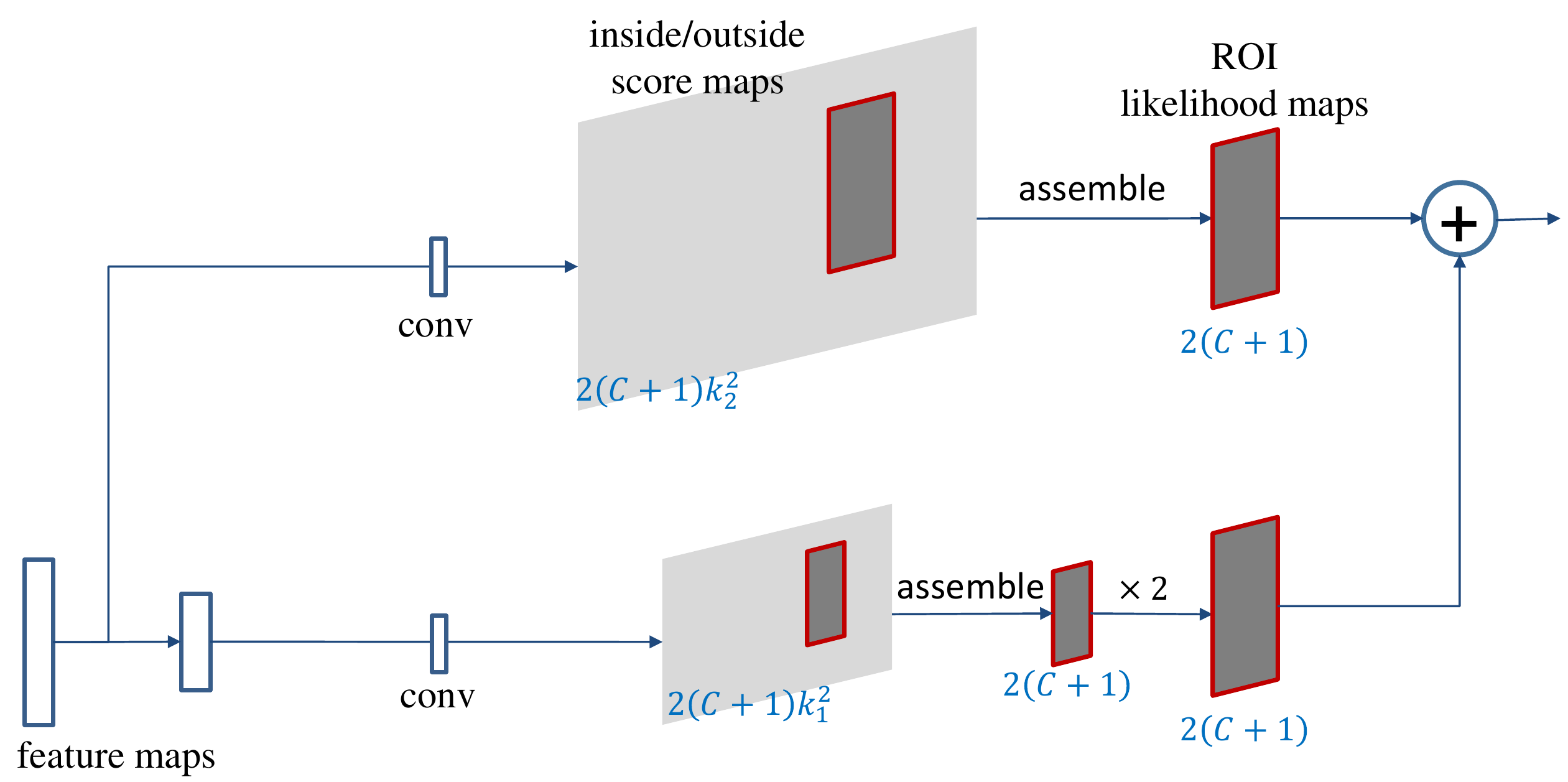}
 \caption{Fusion of multiple score maps at different scales and partition modes. The first set of score maps with $k_1 \times k_1$ partitions and the second set with $k_2 \times k_2$ partitions are generated from conv5 layers and conv3 layers, respectively. Assembled ROI likelihood maps from the first set are $\times 2$ upsampled and added to those from the second set.}
 \label{fig:map-fusion}
\end{figure*}

In our implementation, we generate two sets of position-sensitive score maps from different feature maps. The first set with $2k_1^2 \times (C+1)$ score maps is created from the conv5 layers by a $1 \times 1$ convolutional layer. The second set with $2k_2^2 \times (C+1)$ score maps is created in the same way from conv3 layers, which is $2\times$ larger than conv5 ones. In our experiments, we use $(k_1,k_2)=(7,9)$ by default. Each ROI is projected into a 16$\times$ smaller region over the first set of score maps, and a 8$\times$ smaller region over the second set. Then, for each ROI, we apply the assembling operation on each set of score maps to produce two sets of $2(C+1)$ ROI likelihood maps. In each set of likelihood maps, the first half consists of outside maps, and the second half consists of inside maps. Finally, we upsample the first set of ROI likelihood maps by $\times 2$, and sum the two sets to get the final likelihood maps. The process is illustrated in \figref{map-fusion}.

{\noindent \bf Bayesian Inference:} In the next step, we aim to compute instance segmentation probabilities $I$ for each ROI, which consists of $C + 1$ inside probability maps and $C + 1$ outside probability maps. We first crop the ROI rectangle from the $C+1$ semantic segmentation score maps $S$ to produce $C+1$ ROI semantic segmentation probability maps. We then apply the element-wise product operation on these probability maps and the $C+1$ ROI inside likelihood maps from $L$. The results are $C+1$ ROI inside probability maps, as shown in \figref{biseg-network}. The intuition of the product operation here is that the ROI inside probability maps are predicted as posteriors in Bayesian inference, where the ROI semantic segmentation probability maps are used as priors, and the ROI inside likelihood maps are used as likelihoods. The ROI outside probability maps are equal to the ROI outside likellihood maps because there are no priors to predict outside probabilities. 

To clarify our Bayesian inference formulation, first let $P(I_{ck}|X)$ be the posterior probability of instance segmentation for category $c$ and ROI $k$, given the input image $X$. It can be expressed as:
\begin{equation}
P(I_{ck}|X) = \sum_{\text{category} j}{P(I_{ck}, j|X)}.
\end{equation}
As $P(I_{ck}, j)=0$ for $j \neq c$, we can rewrite that
\begin{equation}
P(I_{ck}|X)=P(I_{ck}, c|X)=P(I_{ck}|c, X)P(c|X).
\end{equation}
Here we use the position-sensitive score map to approximate $P(I_{ck}|c, X)$, and the semantic segmentation score map to approximate the prior $P(c|X)$. As shown in our experiment results, the semantic segmentation priors significantly enhance the instance segmentation inference owing to the strong correlation between the semantic segmentation probabilities and the instance-aware inside likelihoods.

Following FCIS \cite{Li2017}, on top of the inside and outside probability maps, we apply a softmax operation to produce the foreground probability, and a max operation to produce the per-pixel likelihood of the object category, which is then used to infer the classification score by average pooling. 

{\noindent \bf Training:} As in Fast R-CNN \cite{Girshick2015}, an ROI is considered positive if it has Intersection over Union (IoU) with a ground-truth box of at least 0.5 and negative otherwise. We define a multi-task loss as follows:
\begin{equation}
L = L_{rpn} + L_{ss} + L_{cls} + L_{mask} + L_{bbox}.
\label{eq:multitask-loss}
\end{equation} 
The loss of the RPN sub-network $L_{rpn}$ is defined as in Faster R-CNN \cite{Ren2015}. The semantic segmentation loss $L_{ss}$ is per-pixel multinomial cross-entropy loss, as in common FCNs \cite{Long2015}. The last three losses in Eq. \eqnref{multitask-loss} are defined on each ROI: the softmax classification loss $L_{cls}$ over $C + 1$ categories, the binary cross-entropy loss $L_{mask}$ over the foreground mask of the ground-truth category, and the bounding box regression loss $L_{bbox}$ as in Fast R-CNN \cite{Girshick2015}. $L_{mask}$ and $L_{bbox}$ are defined only on positive ROIs.

We use the ImageNet pretrained model \cite{He2016} to initialize the shared convolutional layers. We adopt image-centric training \cite{Girshick2015}: the shared convolutional layers and the semantic segmentation loss are computed on the entire image, while the ROIs are randomly sampled for computing the other losses. In our implementation, each mini-batch involves 1 image, and each image has 64 sampled ROIs. We use SGD optimization. The model is trained using a learning rate of 0.001 for the first 20k, and 0.0001 for the last 10k iterations respectively. We train the model in one GPU, which holds 8 mini-batches.

{\noindent \bf Inference:} At test time, we generate 300 ROIs from RPN. We run the bounding box regression branch on these proposals, followed by non-maximum suppression. The remaining ROIs are classified as the categories with highest classification scores. Following the mask voting scheme in \cite{Dai2016a}, for each remaining ROI, we find its near-by instances which overlap with it by IoU $\ge 0.5$. Their foreground masks of the category are averaged pixel-by-pixel, weighted by their classification scores. The averaged masks are binarized to form the final output masks.

\section{Experiments}
\label{sec:experiment}
Following previous work \cite{Hariharan2014, Hariharan2015, Dai2015, Dai2016a, Dai2016b, Li2017}, we evaluate our instance segmentation performance on the PASCAL VOC dataset \cite{Everingham2010}. We train the model on the VOC 2012 train set (5623 images), and perform evaluation on the VOC 2012 validation set (5732 images)\footnote{There is some confusion about the name of the evaluation dataset. For example, the SBD dataset used in \cite{Arnab2017} is equivalent to the dataset used in our experiments.}, with the additional instance mask annotation from \cite{Hariharan2011}. The evaluation metric we use is the mean Average Precision over regions, which is referred to as mAP$^r$ \cite{Hariharan2014}. We report mAP$^r$ using IoU thresholds at 0.5 and 0.7. Different from the AP metric used in object detection, the IoU is computed over predicted and ground-truth regions instead of bounding-boxes.

Following FCIS \cite{Li2017}, we use ResNet-101 model \cite{He2016} pre-trained on ImageNet \cite{Deng2009}, and do not apply online hard example mining (OHEM) \cite{Shrivastava2016a} on our model. We compare BiSeg with recent methods \cite{Hariharan2014, Hariharan2015, Dai2015, Liu2016, Dai2016a, Dai2016b, Li2016, Arnab2017, Li2017}, and variants of our own method with different settings. The results are shown in \tabref{voc-instanceseg}.

\begin{table}[t]
\begin{center}
\begin{tabular}{l|c|c}
\hline 
method & mAP$^r$@0.5(\%) & mAP$^r$@0.7(\%) \\ 
\hline \hline
SDS \cite{Hariharan2014} & 49.7 & 25.3 \\
Hypercolumn \cite{Hariharan2015} & 60.0 & 40.4 \\
CFM \cite{Dai2015} & 60.7 & 39.6 \\
MPA \cite{Liu2016} & 61.8 & - \\
MNC \cite{Dai2016a} & 63.5 & 41.5 \\
MNC, Instance FCN \cite{Dai2016b} & 61.5 & 43.0 \\
IIS \cite{Li2016} & 63.6 & 43.3 \\
CRF \cite{Arnab2017} & 62.0 & 44.8 \\ 
FCIS \cite{Li2017} & 65.7 & 52.1 \\ 
\hline 
FCIS* & 64.2 & 48.6 \\
naive Multi-task & 65.2 & 49.6 \\
BiSeg (single PS score map) & 66.4 & 50.5 \\
BiSeg (fused PS score map) & {\bf 67.3} & {\bf 54.4} \\
\hline
\end{tabular} 
\end{center}
\caption{Comparisons of instance segmentation on the PASCAL VOC 2012 validation set.}
\label{tab:voc-instanceseg}
\end{table}

{\bf FCIS*:} This baseline is based on our own implementation of FCIS \cite{Li2017} \footnote{The source code of FCIS has not been released at the time of writing this paper.}. It is equivalent to our model that includes only the instance segmentation sub-network.

{\bf naive Multi-task:} To verify the importance of the combination of semantic segmentation and instance segmentation sub-networks, this baseline removes the combination module from the top of BiSeg. It can be considered a common multi-task learning method with two branches sharing the feature map. 

{\bf BiSeg (single PS score map):} This version of our model uses the traditional position-sensitive score maps \cite{Li2017} in the instance segmentation sub-network. The mAP$^r$ scores are 66.4\% and 50.5\% at IoU thresholds of 0.5 and 0.7 respectively. They are 1\% higher than {\bf naive Multi-task}, and 2\% than {\bf FCIS*}. Moreover, our AP$^r$ at 0.5 achieves a 0.7\% improvement compared with the previous state-of-the-art FCIS \cite{Li2017}. This verifies the effectiveness of enhancing instance segmentation by semantic segmentation.

{\bf BiSeg (fused PS score map):} By using the fusion of position-sensitive score maps, we achieve the best results, 67.3\% and 54.4\%, which are nearly 2\% higher than those of FCIS \cite{Li2017}. The AP$^r$ at 0.7 achieves a 4\% improvement compared with {\bf BiSeg (single PS score map)}. This proves the importance of the proposed map fusion approach.

For a deeper understanding on the fusion of position-sensitive score maps, we evaluate BiSeg with different combinations of partition modes $(k_1, k_2)$. As stated in previous work \cite{Dai2016b}, position-sensitive score maps with $7 \times 7$ partitions gave the best results. Therefore, we fix $k_1$ to 7, and test different values of $k_2 \ge 7$. As shown in \tabref{partition_mode}, fusion of score maps with different partition modes $(k_1,k_2)=(7,9)$ is more effective than score maps with the same partition modes $(k_1,k_2)=(7,7)$. However, $k_2$ with a too large value ($k_2=11$) causes a decrease in mAP$^r$@0.5, due to the over-partition of the position-sensitive score maps.

\begin{table}[t]
\begin{center}
\begin{tabular}{c|c|c|c}
$(k_1,k_2)$ & (7, 7) & (7, 9) & (7, 11) \\ 
\hline
mAP$^r$@0.5(\%) & 66.9 & {\bf 67.3} & 66.8 \\
\hline
mAP$^r$@0.7(\%) & 54.1 & {\bf 54.4} & {\bf 54.4} \\
\end{tabular} 
\end{center}
\caption{Comparisons of different combinations of partition modes $(k_1, k_2)$.}
\label{tab:partition_mode}
\end{table}

We also evaluate the performance of semantic segmentation, as shown in \tabref{voc-semseg}. We report here the mean accuracy and the mean region intersection over union (mean IU). Our model outperforms {\bf naive Multi-task}, which learns semantic segmentation and instance segmentation in separated branches. Once again, it verifies the importance of combining the two segmentation tasks in our model.

\begin{table}[t]
\begin{center}
\begin{tabular}{l|c|c}
\hline 
method & mean accuracy(\%) & mean IU(\%) \\ 
\hline \hline
naive Multi-task & 69.0 & 59.5 \\
BiSeg & 70.2 & 60.8 \\
\hline
\end{tabular} 
\end{center}
\caption{Comparisons of semantic segmentation on the PASCAL VOC 2012 validation set.}
\label{tab:voc-semseg}
\end{table}

\begin{figure*}
 \centering
 \includegraphics[width=\textwidth]{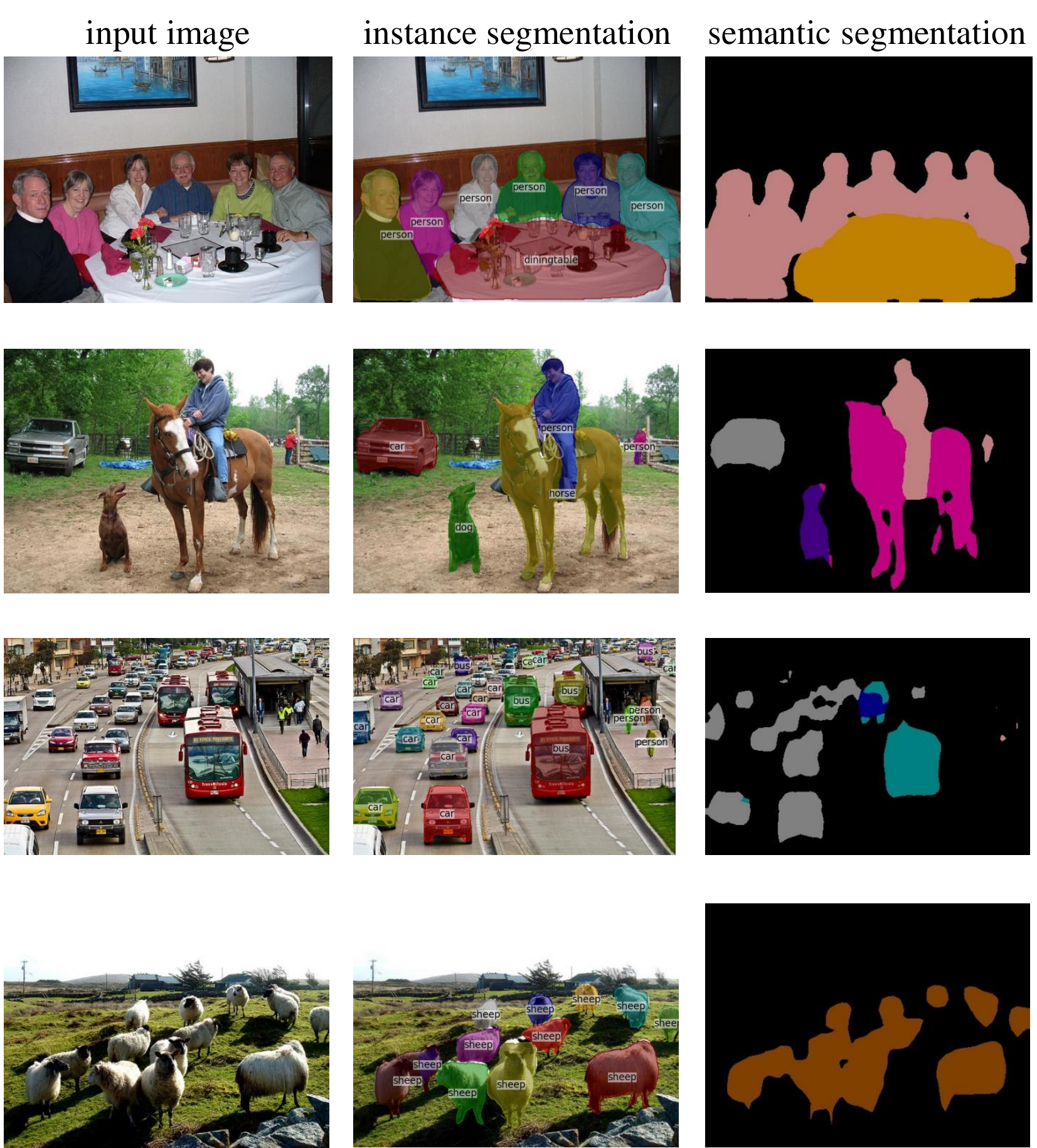}
 \caption{Results of BiSeg. Our method produces high segmentation quality at borderlines without using CRF. For instance segmentation, we show all detected instances whose classification scores are larger than 0.5. We confirm good results of instance segmentation even for heavily occluded scenes (the last two rows).} 
 \label{fig:biseg-results}
\end{figure*}

Finally, we show some of our results for both instance segmentation and semantic segmentation in \figref{biseg-results}. Our method produces high segmentation quality at borderlines without using CRF. Moreover, we confirm good instance segmentation results even for heavily occluded objects. Our approach can be applied to the problem of counting objects from crowded scenes \cite{Pham2015}.

\section{Conclusion}
We present a fully convolutional end-to-end solution for simultaneous semantic segmentation and instance segmentation. Our method predicts the instance segmentation mask as a posterior probability in Bayesian inference, where the semantic segmentation result is treated as prior, and fusion of multiple position-sensitive score maps at different scales and partition modes as likelihood. We achieve the best results, 67.3\% and 54.4\%, on the PASCAL VOC dataset, which are nearly 2\% higher than those of FCIS. In future work, we intend to investigate other variants of context priors such as depth maps and instance-aware likelihoods such as human pose.
\label{sec:conclusion}

\section*{Acknowledgement}
We would like to thank all lab members, in particular Susumu Kubota and Yuta Shirakawa for the helpful suggestions on making this paper better.

\bibliography{ref}
\end{document}